%% file: main.tex
\documentclass{article}
\usepackage[nonatbib,preprint]{neurips}

\usepackage[utf8]{inputenc}
\usepackage[T1]{fontenc}
\usepackage{hyperref}
\hypersetup{
    colorlinks=true,
    citecolor=blue,
    linkcolor=blue,
    filecolor=blue,      
    urlcolor=blue,
    pdftitle={Overleaf Example},
    pdfpagemode=FullScreen,
    }

\usepackage{url}
\usepackage{booktabs}
\usepackage{amsfonts}
\usepackage{nicefrac}
\usepackage{microtype}
\usepackage[table]{xcolor}
\usepackage{ulem}
\usepackage{adjustbox}
\usepackage{multirow}

\usepackage{todonotes}

\usepackage{amsmath}
\usepackage[utf8]{inputenc}
\usepackage[numbers]{natbib}
\usepackage{graphicx}

\usepackage[frozencache,cachedir=.]{minted} 

\setminted{
  frame=none,
  bgcolor=gray!20,
  fontsize=\footnotesize,
  linenos,
  breaklines=true
}

\usepackage{listings}

\definecolor{codebg}{gray}{0.9}
\definecolor{keywordcolor}{rgb}{0.26, 0.35, 0.75}
\definecolor{modulecolor}{rgb}{0.2, 0.3, 0.8}
\definecolor{stringcolor}{rgb}{0.63, 0.12, 0.18}
\definecolor{commentgray}{gray}{0.4}

\lstdefinelanguage{mypython}{
  language=python,
  morekeywords={as, from, import},
  moredelim=[is][\color{modulecolor}]{`}{`},
  alsoletter={.},
}

\usepackage{ulem}

\newcommand{\pnpl}{\includegraphics[height=1.5\fontcharht\font`\B]{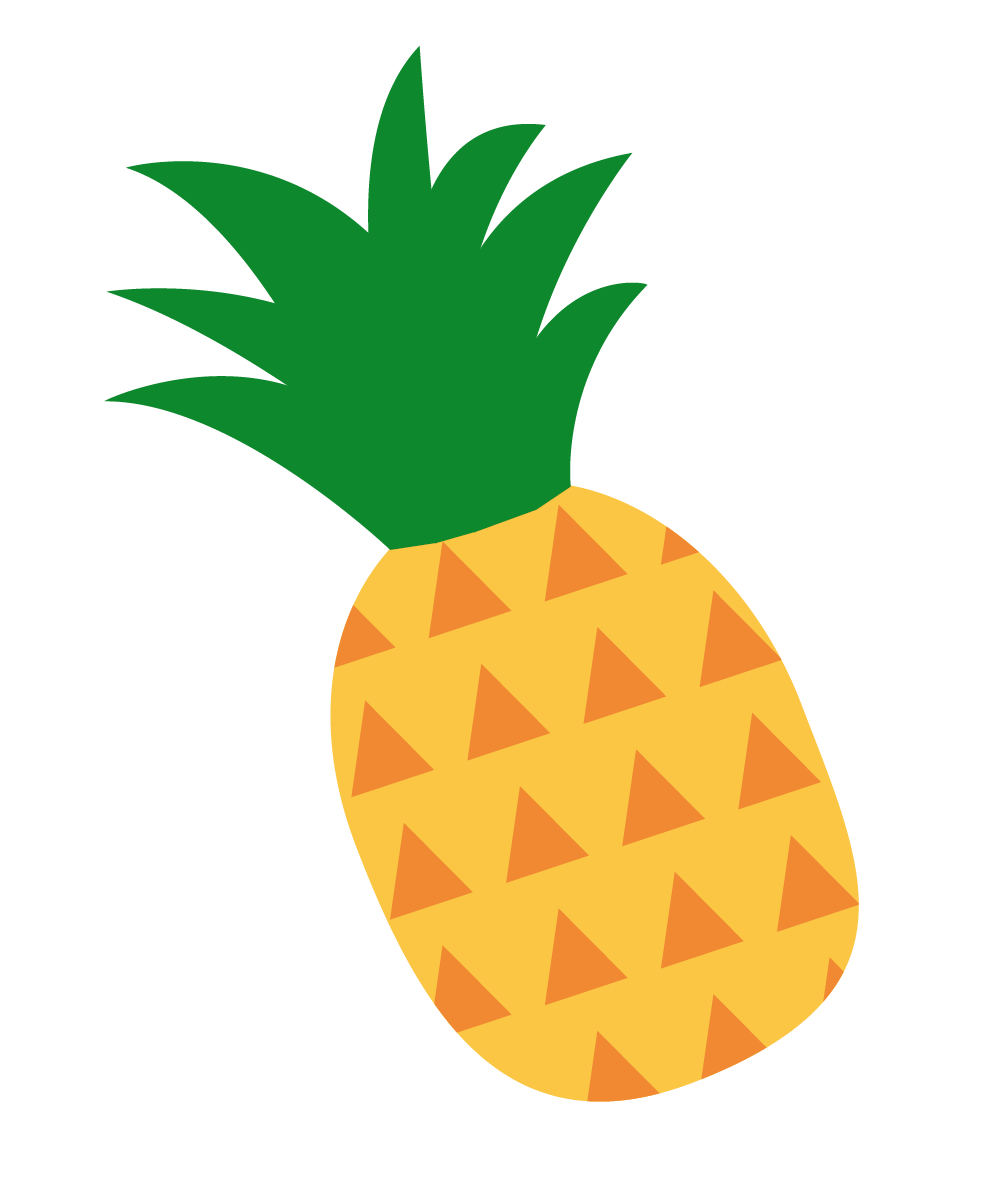}}

\title{The 2025 PNPL Competition: Speech Detection and Phoneme Classification in the LibriBrain Dataset}

\input{sections/0_authors}

\begin{document}

\maketitle

\begin{abstract}

The advance of speech decoding from non-invasive brain data holds the potential for profound societal impact. Among its most promising applications is the restoration of communication to paralysed individuals affected by speech deficits such as dysarthria, without the need for high-risk surgical interventions. The ultimate aim of the 2025 PNPL competition is to produce the conditions for an ``ImageNet moment'' or breakthrough in non-invasive neural decoding, by harnessing the collective power of the machine learning community. 

To facilitate this vision we present the largest within-subject MEG dataset recorded to date (\textit{LibriBrain}) together with a user-friendly Python library (\texttt{pnpl}) for easy data access and integration with deep learning frameworks. For the competition we define two foundational tasks (i.e.~\textit{Speech Detection} and \textit{Phoneme Classification} from brain data), complete with standardised data splits and evaluation metrics, illustrative benchmark models, online tutorial code, a community discussion board, and public leaderboard for submissions. To promote accessibility and participation the competition features a \textit{Standard} track that emphasises algorithmic innovation, as well as an \textit{Extended} track that is expected to reward larger-scale computing, accelerating progress toward a non-invasive brain-computer interface for speech.

\end{abstract}

\subsection*{Keywords}
Neural Decoding, Non-Invasive, MEG, Speech Comprehension, LibriBrain

\section{Competition description}
\input{sections/1-1_background_and_impact}

\input{sections/1-2_novelty}

\input{sections/1-3_data}

\input{sections/1-4_tasks}

\input{sections/1-5_metrics}

\input{sections/1-6_baselines_code_materials}

\input{sections/1-7_website_tutorial_docs}

\section{Organisational aspects}
\input{sections/2-1_protocol}

\input{sections/2-2_rules_and_engagement}

\input{sections/2-3_schedule_and_readiness}

\input{sections/2-4_competition_promotion_and_incentives}

\section{Resources}
\input{sections/3-2_resources_provided}

\input{sections/3-3_support_requested}

\newpage 
\section*{Organising team}
\input{sections/999_team}

\input{sections/999_ack}

\bibliography{references}

\end{document}

%% file: sections/0_authors.tex
\author{
    \textnormal{Gilad Landau\textsuperscript{1,N}} \and
    Miran Özdogan\textsuperscript{1,N} \and
    Gereon Elvers\textsuperscript{1,N} \and
    Francesco Mantegna\textsuperscript{1,N} \and
    Pratik Somaiya\textsuperscript{1,N} \and 
    Dulhan Jayalath\textsuperscript{1,N} \and
    Luisa Kurth\textsuperscript{1,N} \and
    Teyun Kwon\textsuperscript{1,N} \and
    Brendan Shillingford\textsuperscript{2} \and
    Greg Farquhar\textsuperscript{2} \and
    Minqi Jiang\textsuperscript{3} \and
    Karim Jerbi\textsuperscript{4,5} \and
    Hamza Abdelhedi\textsuperscript{4,5} \and
    Yorguin Mantilla Ramos\textsuperscript{4,5} \and
    Caglar Gulcehre\textsuperscript{6} \and
    Mark Woolrich\textsuperscript{7,8,N} \and
    Natalie Voets\textsuperscript{8,N} \and 
    \textnormal{Oiwi Parker Jones\textsuperscript{1,N}} \AND
    \textnormal{\textsuperscript{1}PNPL}\pnpl%
    \and
    \textsuperscript{2}Google DeepMind \and
    \textsuperscript{3}MetaAI \and
    \textsuperscript{4}Mila \and
    \textsuperscript{5}Université de Montréal \and
    \textsuperscript{6}EPFL \and
    \textsuperscript{7}OHBA \and
    \textsuperscript{8}WIN \and
    \textsuperscript{N}University of Oxford \AND
{\tt \{gilad, oiwi\}@robots.ox.ac.uk}}

%% file: sections/1-1_background_and_impact.tex
\subsection{Background and impact}

\input{figures/fig_teaser}

Until a few years ago, the idea of communicating through a neural prosthetic was confined to science fiction. 
Today the frontier of neuroscience and artificial intelligence (AI) holds the potential of expansive benefits for people who have lost the ability to speak because of debilitating ailments, such as traumatic brain injury and motor neurone disease. 
In 2021, the first brain-computer interface (BCI) capable of restoring connected-speech to a paralysed individual appeared \citep{moses2021nejm}. 
This landmark BCI enabled sentence decoding from surgically-implanted electrodes for a limited 50-word vocabulary, achieving a promising 25.6\% word-error rate (WER).

Since then, new surgical data and AI systems have resulted in BCIs with vocabulary sizes increasing up to 125,000 words \citep{willett2023high, card2024nejm} and reported WERs decreasing down to less than 5\% \citep{card2024nejm}. 
This is remarkable as 10\% WER is often cited as the threshold for widespread adoption of automatic speech recognition (ASR), when it reached human parity \citep{xiong2016parity} on realistic conversational 
benchmarks  \citep{godfrey1992switchboard}. 

Despite this rapid progress, invasive BCIs face fundamental limitations to widespread adoption. Two critical limitations are that brain surgery is inherently dangerous and that surgical data does not easily scale. By contrast, non-invasive neuroimaging offers a safe alternative that can be easily repeated in healthy participants, unlocking data collection at unprecedented scale. 
Amongst current technologies, magnetoencephalography (MEG) stands out for its unique strengths. 
It is a direct measure of neuronal activity, providing millisecond temporal resolution equivalent to intracranial recordings. 
In this respect, MEG is also like electroencephalography (EEG). 
But, whereas the spatial resolution of EEG is significantly degraded by volume conduction and electrical artefacts from the skull and scalp, the magnetic fields that MEG measures pass through biological tissue without distortion. 
MEG systems routinely achieve spatial localisation in the range of 5--10 mm \citep{vanes2025osl-ephys}, and, in some studies, resolution as precise as 2 mm has been achieved \citep{barratt2018meg2mm}. 
MEG recordings can therefore approach, or, under ideal conditions, even exceed the precision of invasive modalities for speech BCIs such as electrocorticography (ECoG) \citep{moses2021nejm, metzger2023neuroprosthesis}. 

Compared to invasive modalities, a key limitation of MEG is the greater distance between sensors and neural sources, which typically results in lower signal-to-noise ratios. 
Our working hypothesis is that with enough high-quality data, and with the right deep learning methods, MEG-based speech decoding should be able to compete with surgical alternatives. 
We are therefore proposing this competition to coincide with PNPL's first big release of data. 
Prior to this, the biggest public EEG and MEG datasets have typically included tens to hundreds of participants, but only 1--2 hours per person \citep{nieuwland2018, schoffelen2019, broderick2019, dascoli2024, accou2023, gwilliams2023megmasc}, resulting in data that can be characterised as \textit{broad} but \textit{shallow}. 
An exception is a dataset including 10 hours of within-subject MEG data \citep{armeni2022}. 
Empirical results show that \textit{deep} data, representing big data acquired over repeated scans from the same subject, yield the largest gains in decoding performance \citep{dascoli2024}. 
So, it is good timing for us to be releasing the \textit{deepest} within-subject, speech decoding dataset to date. 
In this first release of the LibriBrain dataset \citep{ozdogan2025libribrain}, we include over 50 hours of MEG, which is 5$\times$ the previously biggest dataset \citep{armeni2022}, and 25--50$\times$ bigger than most other MEG and EEG datasets. 
Given that data is king, we expect the competition to break new ground in non-invasive speech decoding (see Figure \ref{fig_teaser}).

The first PNPL competition will focus on two basic but fundamental tasks, which have been influential in the development of both ASR and invasive brain-to-text (B2T). 
Several recent efforts to train non-invasive B2T systems with MEG or EEG have reported WERs approaching 100\%, indicating uninformative or near-chance performance \citep{jo2024eegtotext, yang2024neuspeech, yang2024mad, yang2024neugpt}. 
Other recent non-invasive works have found success by exploring a diverse set of simpler tasks such as Segment Identification \citep{defossez2023}, Word Classification \citep{dascoli2024}, and (Phonetic) Feature Classification \citep{jayalath2024scaling}. 
To make the most of the available data, the 2025 PNPL competition will focus on the tasks of Speech Detection and Phoneme Classification. 
To illustrate the efficiency of these tasks, consider that the LibriBrain dataset includes 1,523,920 phoneme examples divided over 39 phoneme classes. 
Compare this to a task like Word Classification, where an \textit{order of magnitude} fewer examples (466,264 words) are divided over many more classes (e.g.~16,892 distinct words). Even if one were to limit the vocabulary to the most frequent 250 words \citep{dascoli2024}, or even to the most frequent 50 words \citep{moses2021nejm}, Word Classification leverages fewer examples per class. Of course, restricting the vocabulary also results in excluded data. 
Speech Detection is even more efficient as we make a binary prediction for every temporal sample (250 per second). 
Inspired by the success of ImageNet \citep{russakovsky2015imagenet}, our vision is to repeat the PNPL Competition over multiple years. So, as the community solves foundational tasks like Speech Detection and Phoneme Classification, and as our dataset grows ever bigger, we will host a series of future PNPL Competitions across a curriculum of increasingly difficult tasks. 

To encourage both advances in the state-of-the-art and inclusion of participants in the PNPL competition, we will host two tracks for each task. 
In the Standard track, participants will train and test their submissions on data from the LibriBrain dataset. 
This track aims to make the competition relevant and accessible to all participants by levelling the playing field in terms of training data; we expect that this track will reward methodological innovations. 
In the Extended track, participants can train their submissions on any data they want. Recent work has shown that MEG data from multiple datasets can be effectively pooled to improve downstream decoding performance with unsupervised pre-training \citep{jayalath2024scaling}, with domain adaptation \citep{ridge2024domainshift}, and by selectively combining datasets based on a measure of their quality \citep{jayalath2025cracking}. 
To enable useful comparisons, teams competing on the Extended track will evaluate their models on the standard LibriBrain holdout splits. This track is meant to encourage the use of more compute, in order to see how far teams with resources can push the state-of-the-art. 

The 2025 PNPL competition is a collaborative effort by researchers from academia and industry, and from a number of places around the world. 
To make the competition as accessible as benchmark tasks in computer vision, like CIFAR-10 \citep{krizhevsky2009cifar10}, we have put a lot of effort into the materials supporting the competition. 
Chief amongst these is a Python library, which automatically downloads the data as needed and makes it straightforward to integrate into deep learning frameworks like PyTorch \citep{paszke2019pytorch}. 
To illustrate how easy it is to use, note that the library can be installed with one line on the command line (\texttt{pip install pnpl}). The dataset is then accessed like other popular datasets (e.g.~torchvision and huggingface) in Python (e.g.~\texttt{from pnpl.datasets import LibriBrain}). The data is structured into PyTorch-ready tensors, making it easy to integrate with existing machine learning pipelines. The library also comes with methods to generate TSV files for predictions on the competition holdout data, which in turn can be uploaded on a submission website to populate Papers-With-Code style leaderboard plots -- gamifying the experience, we hope, in a fun and motivating way for participants.

%% file: figures/fig_teaser.tex
\begin{figure}[ht]
    \centering
\includegraphics[width=\linewidth]{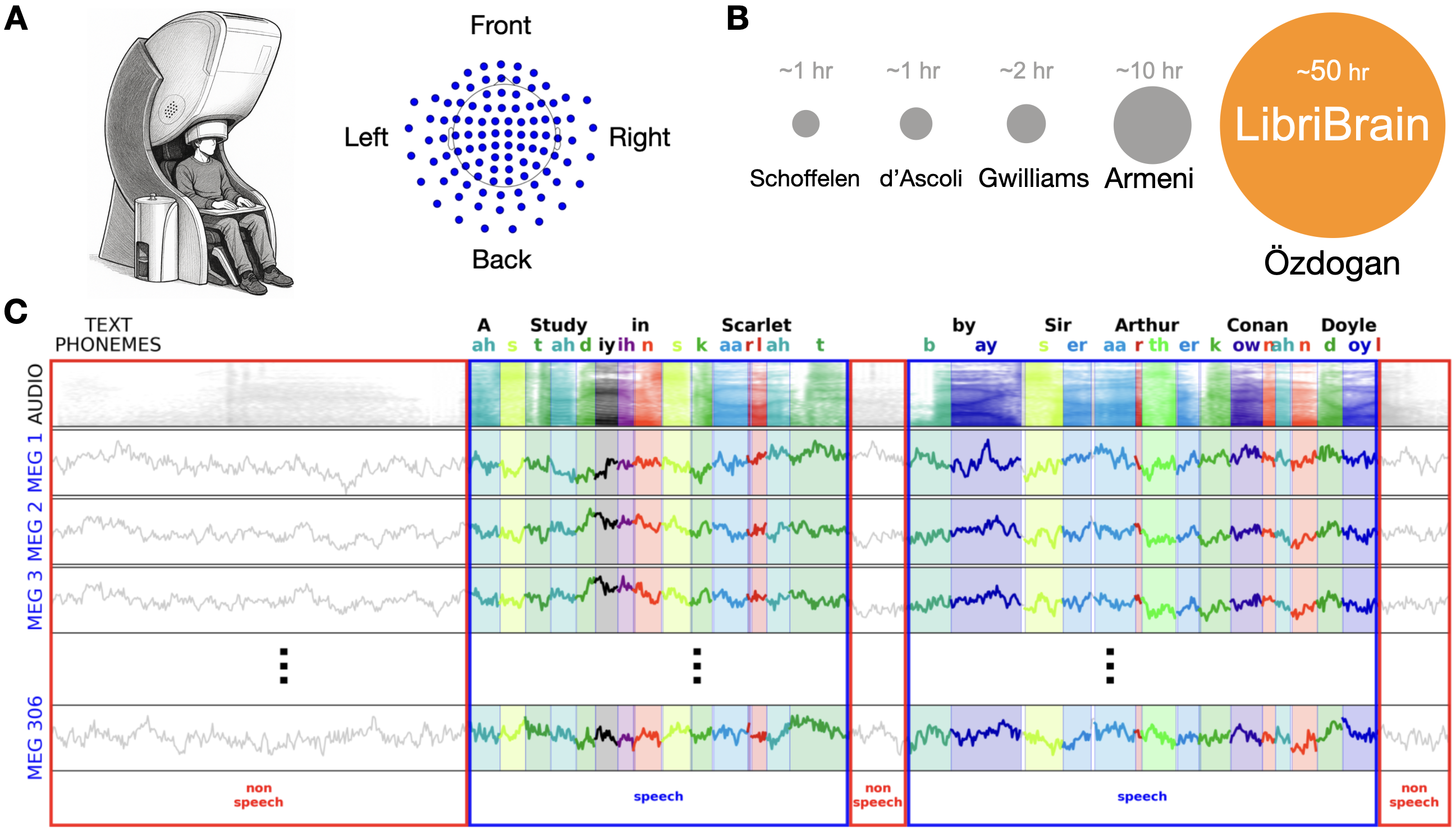}
    \caption{Overview of data and labels. (A) Illustration of a non-invasive MEG scanner and corresponding sensor layout. The layout shows a 2D top-down projection of the MEG sensor positions on the scalp. (B) Schematic comparing the per-subject recording durations of comparable MEG datasets (gray) \citep{schoffelen2019, dascoli2024, gwilliams2023megmasc, armeni2022} with the LibriBrain dataset (orange) \citep{ozdogan2025libribrain}. LibriBrain is approximately 5$\times$ bigger than the next biggest dataset, and 25-50$\times$ bigger than typical datasets. (C) Example MEG recordings with annotations, showing aligned audio, phoneme labels, and speech/non-speech segments. These labels provide the basis for the Speech Detection and Phoneme Classification tasks.}
    \label{fig_teaser}
\end{figure}

%% file: sections/1-2_novelty.tex
\subsection{Novelty}

This is the first competition dedicated to language decoding from non-invasive brain data. An invasive B2T competition was run in 2024 \citep{willett2024benchmark} which helped to bridge the jump from a large-vocabulary WER of 23.8\% \citep{willett2023high} to less than 5\% \citep{card2024nejm}. 
Recently, the non-invasive field has seen remarkable advances, including the availability of high-quality brain recordings and application of powerful new AI methods \citep{defossez2023, dascoli2024, jayalath2024scaling}. 
Yet, despite this progress, we believe that several key elements are still missing—and that addressing them will be essential to unlocking the next major breakthrough. This competition is designed to help close these gaps by focusing on: a the largest dataset of its kind; standard data splits to encourage comparable results in the literature; foundational tasks and evaluation metrics; a competition website; a Python library that allows easy downloading and streamlined integration into deep learning frameworks; tutorial code that runs in the browser with a free GPU for easy onboarding; baseline models that illustrate the viability of the tasks; interactive leaderboards for each competition track; a community Discord for discussion with the organisers and others; and incentives including at least \$10,000 in prizes.

%% file: sections/1-3_data.tex
\subsection{Data}\label{data}

The LibriBrain data used for the competition \citep{ozdogan2025libribrain} are non-invasive MEG recordings acquired from one healthy participant listening to over 50 hours of audiobooks, all sourced from LibriVox \citep{librivox}. 
The MEG recordings were acquired from 306 sensors covering the whole head/brain. %
Neural data were minimally filtered (e.g.~to remove line noise and drift) and downsampled to 250 Hz (see the dataset paper for details \citep{ozdogan2025libribrain}). 
From the perspective of the competition, the MEG data can be thought of as 2-dimensional tensors composed of sensors $\times$ temporal samples (see Figure \ref{fig_teaser}C). 
LibriBrain comes with word- and phoneme-level alignments, as in the LibriSpeech corpus \citep{panayotov2015librispeech}, itself a standard resource for automatic speech recognition (ASR). 
Data and labels can be easily batched using the PNPL library. In practice, one may install the PNPL library from the command line:
\begin{minted}{bash}
pip install pnpl
\end{minted}
The following is a minimal example that illustrates how only a few lines of Python are required to download the data to a specified folder on the user's computer (e.g.~\texttt{/data}):
\begin{minted}{python}
from pnpl.datasets import LibriBrainSpeech
_ = LibriBrainSpeech(data_path="/data")
)\end{minted}

If not already present, the dataset will be automatically downloaded to the specified directory. Alternatively, the data may be downloaded manually from Hugging Face.\footnote{\url{https://huggingface.co/datasets/pnpl/LibriBrain}} 
MEG data and labels are saved in HDF5 and TSV files, respectively. 
The full, serialised LibriBrain dataset is approximately 50 GB. 
The data are standardly split into train, validation, and test sets. For the competition, we include an additional competition holdout split, which includes disjoint subsets of data to be used to update the leaderboard during the competition and to decide the final ranking of submissions (see Table \ref{tab_data-splits}).

\input{tables/tab_data-splits}

%% file: tables/tab_data-splits.tex
\begin{table}[t]
\centering
\small %
\caption{Overview of data splits and approximate durations. The competition holdout set consists of two disjoint subsets used for leaderboard updates and final rankings.}
\label{tab_data-splits}
\begin{tabular}{lcccccc}
\toprule
& \multirow{2}{*}{\textbf{Train}} & \multirow{2}{*}{\textbf{Validation}} & \multirow{2}{*}{\textbf{Test}} & \multicolumn{2}{c}{\textbf{Competition Holdout}} \\
\cmidrule(lr){5-6}
& & & & {Leaderboard} & {Final Rankings} \\
\midrule
\textbf{Data} &
  \cellcolor{green!10}Public &
  \cellcolor{green!10}Public &
  \cellcolor{green!10}Public &
  \cellcolor{green!10}Public &
  \cellcolor{green!10}Public \\
\textbf{Labels} &
  \cellcolor{green!10}Public &
  \cellcolor{green!10}Public &
  \cellcolor{green!10}Public &
  \cellcolor{red!10}Private &
  \cellcolor{red!10}Private \\
\textbf{Hours} &
  51.57 &
  0.36 &
  0.38 &
  \cellcolor{red!10}Private &
  \cellcolor{red!10}Private \\
\bottomrule
\end{tabular}
\end{table}

%% file: sections/1-4_tasks.tex
\subsection{ Tasks and application scenarios}

The tasks in the contest are foundational to neural decoding: Speech Detection and Phoneme Classification. 
Both are supervised and have analogues in the acoustic domain, where early ASR systems were built around phonemes \citep{lee1989speaker, garofolo1993timit, rabiner1993fundamentals} and speech detection has long been a useful preprocessing step in the ASR pipeline \citep{furui2000speech, benesty2007speech}. %
Although we would ultimately like to decode connected speech from the brain, early attemps in EEG and MEG have yielded weak results \citep{yang2024neuspeech, yang2024mad, yang2024neugpt}. There are, however, useful application scenarios for Speech Detection and Phoneme Classification, given their roles in ASR. 
Historically, speech detection played a critical role in the pipeline of the first invasive speech BCI for a paralysed individual \citep{moses2021nejm}: in this work the user attempted to produce isolated words, so that speech detection also functioned as word detection, and word classification together with an external language model were used to produce transcriptions. 
Before this work on decoding full transcripts, many invasive studies focused on phoneme classification \citep{blakely2008localization, guenther2009wireless, leuthardt2011using, pei2011, brumberg2011phonemeclassification, wilson2020allphonemes}. 

When detecting speech from MEG, the input data $x$ is a 306 sensor $\times$ $T$ samples tensor. For each sample $t \in \{ 1, \dots, T \}$ we have a corresponding label $y$, where $y=1$ if speech or $y=0$ if non-speech. %
For the phoneme classification task, the dataloader provides fixed windows of data. Each window is an input $x$ with an associated label $y$, which is one of the standard 39 phoneme classes in the CMU/ARPAbet \citep{cmudict}. %
The task for each input is to predict which class the brain data corresponds to. So rather than \textit{where}, as is emphasised in the Speech Detection task, the Phoneme Classification task is primarily about \textit{what} examples are.

%% file: sections/1-5_metrics.tex
\subsection{Metrics}\label{sec:metrics}

The metric that we use to measure success in both tasks is the F1-macro score:
\[
\mathrm{F1}_{\text{macro}} = \frac{1}{K} \sum_{k=1}^{K} 2 \cdot \frac{\text{Precision}_k \cdot \text{Recall}_k}{\text{Precision}_k + \text{Recall}_k}
\]
This is the unweighted average of per-class F1 scores, where the F1 score is the harmonic mean of Precision and Recall. Here, each class $k$ is a phoneme for Phoneme Classification, or it is speech or non-speech for Speech Detection. For reference, Precision and Recall can be defined in terms of True Positives (TP), False Positives (FP), and False Negatives (FN), each of which is an integer count:
\[
\mathrm{Precision} = \frac{\mathrm{TP}}{\mathrm{TP} + \mathrm{FP}}, \quad
\mathrm{Recall} = \frac{\mathrm{TP}}{\mathrm{TP} + \mathrm{FN}}, \quad \mathrm{F1} = \frac{2 \cdot \mathrm{TP}}{2 \cdot \mathrm{TP} + \mathrm{FP} + \mathrm{FN}}
\]

In practice $\mathrm{F1}_{\text{macro}}$ scores range from 0 to 1 and can be reported as percentages, which we prepose to do in order to communicate proportions clearly in the competition. For Speech Detection, random guessing under balanced-class assumptions yields $\mathrm{F1}_{\text{macro}} = \frac{1}{2} = 50\%$. It is possible, however, to do worse than this in imbalanced settings, since $\mathrm{F1}_{\text{macro}}$ penalises poor performance on the minority class. For example, a naive baseline that always predicts the most frequent label in the LibriBrain data results in a comparatively lower $\mathrm{F1}_{\text{macro}}$ score of 45.3\%. 
For Phoneme Classification with 39 balanced classes, the chance of guessing the correct label is $\frac{1}{39} \approx 2.56\%$. Higher $\mathrm{F1}_{\text{macro}}$ scores are better. 
For the competition, a good predictive model should produce Speech Detection scores significantly greater than 50\%, and Phoneme Classification scores greater than 2.56\% (cf.~Table \ref{tab_baseline-results}).

%% file: sections/1-6_baselines_code_materials.tex
\subsection{Baselines, code, and material provided}

To help participants get started easily, we have created a Starter Kit that includes a Python library for easy data access, clear tutorials on training baseline models, and baseline scores for comparison. 
The first component of the starter kit is the PNPL library introduced in section \ref{data}. It simplifies loading train, validation, and test data and integrating them into minimal training and evaluation loops (similar methods are included to load the competition holdout data and produce submission files):
\begin{minted}{python}
from pnpl.datasets import LibriBrainSpeech
from torch.utils.data import DataLoader

train_data = LibriBrainSpeech(data_path="/data", partition="train")
train_loader = DataLoader(train_data, batch_size=32, shuffle=True)
for x, y in train_loader:    # Iterate through DataLoader as usual
    pass                     # Perform operations using data x and labels y
\end{minted}

As an entry point to the library, we provide three tutorials in the form of Jupyter Notebooks, optimised for usage in Google Colab.
\begin{enumerate}
    \item The first tutorial covers data loading and provides foundational information on the problem domain. It also contains the code for training our reference speech detection model. It is suitable for beginners with some previous familiarity in machine learning without any knowledge in the neuroscience domain.
    \item The second tutorial covers the phoneme classification task in more detail. It assumes familiarity with the dataset and is primarily intended to provide an introduction to the second track. It also contains the code for training our reference phoneme classification model.
    \item The third tutorial explains how to run and submit model predictions, to populate the leaderboards and participate in the competition.
\end{enumerate}

The architectures of the reference models in the tutorials are designed to significantly outperform naive baselines while also being simple enough for beginners to pick up and train on limited amounts of data within the GPU compute provided. We note that the Google Colab Free Tier provides users with a free GPU to use in the browser\footnote{\url{https://research.google.com/colaboratory/faq.html\#gpu-availability}}. The tutorial code has also been tested on Unix (e.g.~Mac) and Windows machines. 

For the speech detection task, we provide a fully trained reference model and report its performance using the evaluation metric ($\text{F}1_{\text{macro}} = 68.04\%$). To ensure meaningful evaluation, we compare the reference model to a naive strategy that always predicts the majority class ($\text{F}1_{\text{macro}} = 45.30\%$). We chose a majority class baseline as the evaluation data were collected using a naturalistic experimental design and contain imbalanced classes---mostly speech with short intermittent periods of non-speech. 
For the phoneme classification task, we provide a reference model for predicting %
phonemes averaged over sets of 100 samples ($\text{F}1_{\text{macro}} = 60.39\%$). We compare this against a naive baseline that always predicts the most likely phoneme (majority class) in the training data ($\text{F}1_{\text{macro}} = 0.47\%$). 
See Table \ref{tab_baseline-results} for a summary.

\input{tables/tab_baseline-results}

%% file: tables/tab_baseline-results.tex
\begin{table}[t]
\centering
\small
\caption{F1-macro scores (\% $\uparrow$ higher is better) for reference models and naive baselines.}
\label{tab_baseline-results}
\begin{tabular}{lrr}
\toprule
\textbf{Method} & \textbf{Speech Detection} & \textbf{Phoneme Classification} \\
\midrule
Reference Model     & \textbf{68.04}\% & \textbf{60.39}\% \\ %
Naive Baseline     & 45.30\% & 0.47\% \\
\bottomrule
\end{tabular}
\end{table}

%% file: sections/1-7_website_tutorial_docs.tex
\subsection{Website, tutorial and documentation}

The website for the competition\footnote{\url{https://neural-processing-lab.github.io/2025-libribrain-competition/}} serves as a hub, aggregating a Starter Kit of code, documentation, tutorials, leaderboards for tracking submissions, and links to the submission system. 
To maximise ease of onboarding for participants, we have produced tutorials for all aspects of the competition in the form of interactive Colab notebooks (e.g.~data exploration, task-specific models, submission). 
To encourage community building, participants also have access to a custom Discord server\footnote{\url{https://neural-processing-lab.github.io/2025-libribrain-competition/links/discord}} where official announcements will be posted and where participants will be encouraged to discuss their approaches, team up, and exchange ideas. Organisers will be available on Discord. We are also planning a series of blog posts to release at regular intervals during the competition to help advertise the competition, and to inspire dialogue within the community.

%% file: sections/2-1_protocol.tex
\subsection{Protocol}

Inclusivity is central to our competition. As such, participating is designed to be simple and accessible. For instance, we have ensured that participants can make impactful contributions without expensive hardware or significant technical barriers. To participate, one needs only to install a Python library (\texttt{pip install pnpl}), which, when used, automatically downloads the dataset and provides seamless integration via a standard PyTorch Data Loader. Once the data is loaded, participants are free to develop their solutions and make predictions on the competition holdout data. The PNPL library includes a straightforward method to write these predictions to a TSV file. Submissions are made by manually uploading TSV files to the EvalAI platform. Solutions will be evaluated automatically according to the metrics detailed in section \ref{sec:metrics}. Rankings for each submission are continuously updated and displayed on our leaderboard. Comprehensive details for every stage of participation, along with tutorials and Colab notebooks equipped with free GPU access, are available on the competition website.

The challenge will consist of two phases for the two tasks. In the first phase, we will release the competition holdout data for the Speech Detection task. Participants will then be able to make official submissions to ``Standard'' and ``Extended'' tracks. The second phase will commence after the Speech Detection tracks have closed. In this phase, we will release the competition holdout data for the Phoneme Classification task, and participants will be invited to make submissions to two new tracks.

%% file: sections/2-2_rules_and_engagement.tex
\subsection{Rules and engagement}

\begin{itemize}
    \item This challenge encourages everyone to join and helps us bring forth the societal impact of speech decoding technologies.
    \item No domain specific knowledge or specialised hardware is required (free GPU access is available through Google Colab).
    \item Although our goal is to democratise the field of speech decoding, we are aware that previous breakthroughs in AI were the result of accumulating more data and computational resources. To resolve this tension, we are splitting the competition into two tracks for each task.
    \item In the ``Standard'' tracks participants may only use the LibriBrain data in their solutions, empowering teams with fewer resources to compete by innovating on the methods side. 
    \item Pre-trained models may be used, provided they were publicly available and known to the community prior to the launch of the competition (e.g.~described in a NeurIPS paper and downloadable from Hugging Face). The organisers reserve the right to make final decisions on any edge cases. Think of this as a hinge loss: staying away from the margin will be safer.
    \item In the ``Extended'' tracks, there are no limits to the training data that participants can use.
    \item Teams may submit to all tracks and their progress will be shared on the relevant leaderboards. However, to encourage diversity, any team will only be allowed to win prize money for one track. If the same team were to win multiple tracks, then they would still be listed on the leaderboards, but the next team in the ranking would get the smaller of the prizes. 
    \item The organisers strongly encourage all participants to share their code in the spirit of open-science. Similarly, participants are invited to submit pull requests to the GitHub repo for the PNPL library (e.g.~to add dataloaders to new datasets and thus accelerate the community).
    \item The organisers will reach out to participants on the ``Standard'' tracks who could win prizes to verify that their solutions are not dependent on external datasets by sharing training code, and to participants on the ``Extended'' tracks to query the amount of compute used. For verification, this may also require models or code by request after the relevant closing date. %
    \item To avoid data leakage, final ranking will be determined based on independent holdout data. 
    \item The top 3 confirmed submissions for each track will win, provided they beat the reference models in Table \ref{tab_baseline-results} for each task (i.e.~F1$_{\text{macro}}$ scores of 68.04\% for Speech Detection and 60.39\% for Phoneme Classification). In the unlikely event of a tie, the prize will be split.
\end{itemize}    
To facilitate open and continuous communication between the organisers and participants, a dedicated
Discord channel will be used as the primary platform for all contest-related discussions. This includes
addressing specific questions, enabling real-time discussion, and providing technical support. The
forum will be managed collaboratively by contest organisers to ensure comprehensive support.

%% file: sections/2-3_schedule_and_readiness.tex
\subsection{Schedule and readiness}

The organisers are ready at the time of submission with working versions of all competition materials. 

The proposed schedule will commence advertisement on the day that competition acceptances are announced. We will then provide a few weeks for participants to familiarise themselves with the website and with restricted beta versions of the competition materials. To promote a fair competition, the full set of materials will be released for the Speech Detection Tracks on 1 June, 2025, with 2 months allocated for submissions. The Phoneme Classification Tracks will then run for another 2 months, leaving time for the organisers to analyse the results and prepare analyses to present at our NeurIPS session.  By staggering the tasks, we hope to generate renewed excitement within the community for the Phoneme Classification Tracks when we announce results from the Speech Detection Tracks. This will also give the organisers more time to verify the Speech Detection results, establishing a protocol to turn around the Phoneme Classification results more quickly.

\textbf{\sout{10 May} 26 May, 2025}: Acceptance notifications for competition proposals are sent out. The official contest announcement and promotion commence. Beta versions of all necessary resources will be made available through the competition website so participants can familiarise themselves with the tutorial code and set up their training environments. This time will allow participants to familiarise themselves with the materials and seek clarification from the organisers.

\textbf{1 Jun -- 31 Jul, 2025}: Official opening of the contest to the public. Phase 1 commences with the start of the Standard and Extended Speech Detection Tracks. All relevant materials released.

\textbf{1 Aug -- 30 Sep, 2025}: Phase 2 begins with the release of all materials for the two Phoneme Classification Tracks (Standard and Extended). During this phase, the organisers will evaluate submissions from phase 1, with contenders for prizes being asked to share system details (e.g.~training code for the Standard Tracks) to ensure winners adhere to the rules of the competition. 

\textbf{1 Oct -- 10 Oct, 2025}: The organising committee reviews and verifies all results. Winners of the first two Speech Detection Tracks will be announced on the competition website. Contenders for prizes in the second two Phoneme Classification Tracks will be asked to share details about their submissions with the organisers. The organisers will endeavour to contact winners before the \href{https://neurips.cc/Conferences/2025/Dates}{NeurIPS Early Registration Deadline} (11 Oct, 2025), noting the \href{https://neurips.cc/FAQ/CancellationPolicy}{Visa Application Deadline} (16 Oct, 2025).

\textbf{15 Oct -- 1 Dec, 2025}: Organisers conduct an in-depth analysis of contest results for the conference.

\textbf{2 Dec -- 7 Dec, 2025}: The contest will culminate in a dedicated session at NeurIPS. Winners will be announced, prizes will be awarded, and select participants will be invited to present. 

At the time of writing this proposal, all of the materials for the competition are ready but have not yet been made publicly available. The equivalent of \$10,000 USD has been raised in GBP for prizes. Additional funds are being sought to be able to offer travel and computational support to participants.

%% file: sections/2-4_competition_promotion_and_incentives.tex
\subsection{Competition promotion and incentives}

We will promote the competition vigorously through social networks, academic mailing lists, and other venues. Special attention will be taken to reach out to all listed NeurIPS affinity groups (e.g.~Indigenous in AI, LatinX in AI, Black in AI, Women in ML). 
A dedicated website has been created, providing information, resources, and continuous updates about the competition. 
During the competition, the organisers plan to release a series of blog posts that go in-depth into relevant questions from what we know about the tasks, with the aim of creating discussion and further interest in the competition. 
Participation will be incentivised through rewards, with at least \textbf{\$10,000} available for prizes. 
To promote inclusivity, we are in ongoing discussions to raise additional funds for travel and compute, especially for participants from under-represented groups in AI and ML. %
We intend to invite teams with innovative solutions to participate in a joint publication.

%% file: sections/3-2_resources_provided.tex
\subsection{Resources provided by organisers}

The organisers will actively monitor and support the competition, offering technical and scientific guidance through Discord and the competition website. 
Tutorial code can run in the browser and provides a free GPU for participants. A Python library is available to download the data and integrate it into standard deep learning frameworks. 
Winners will receive prize money. %

%% file: sections/3-3_support_requested.tex
\subsection{Support requested}

The main support needed will be the provision of a video-conferencing platform/setting for presentations, and a room for in-person participant gathering and presentations. 
If possible, we would like to request registration be waived for up to one member apiece from the top-3 submissions in each track, as well as 3 discretionary places (e.g.~from under-represented groups; so 15 participants total).

%% file: sections/999_team.tex
Our team combines AI researchers and neuroscientists, and embodies a strategic collaboration between researchs in Oxford, Montreal, Switzerland, Google, and Meta. We range in levels of seniority from research assistants and master's students, to DPhil and PhD students, Professors and Research Scientists in Industry. 
In terms of diversity, our team includes members of multiple NeurIPS affinity groups (e.g.~Indigenous in AI and Women in ML).

\textbf{Gilad Landau} is a DPhil student in Engineering and member of PNPL{\pnpl}, supervised by Oiwi Parker Jones at the University of Oxford. 

\textbf{Miran Özdogan} is a DPhil student in Computer Science and member of PNPL{\pnpl}, supervised by Oiwi Parker Jones and Michael Bronstein, at the University of Oxford. 

\textbf{Gereon Elvers} is a master's student in Information Systems at the Technical University of Munich (TUM) and visiting researcher at PNPL{\pnpl}. 

\textbf{Francesco Mantegna} is a Postdoctoral Researcher in PNPL{\pnpl}, Department of Engineering Science, University of Oxford. He received his PhD from NYU under the supervision of David Poeppel. 

\textbf{Pratik Somaiya} is a Software Engineer at the Oxford Robotics Institute. 

\textbf{Dulhan Jayalath} is a DPhil student in Machine Learning at the University of Oxford, supervised by Oiwi Parker Jones as part of PNPL{\pnpl} and funded by an Amazon Web Services (AWS) studentship as part of the EPSRC Centre for Doctoral Training in Autonomous Intelligent Machines and Systems (AIMS). 
His research interests lie in a range of areas including brain decoding, LLMs, reasoning, and foundation models. 

\textbf{Luisa Kurth} is a DPhil student in PNPL{\pnpl} and the EPSRC Centre for Doctoral Training in Autonomous Intelligent Machines and Systems (AIMS), University of Oxford.

\textbf{Teyun Kwon} is a DPhil student in PNPL{\pnpl} and the EPSRC Centre for Doctoral Training in Autonomous Intelligent Machines and Systems (AIMS), University of Oxford.

\textbf{Brendan Shillingford} is a Staff Research Scientist at Google DeepMind. 

\textbf{Greg Farquhar} is a Staff Research Scientist at Google DeepMind. %

\textbf{Minqi Jiang} is a Senior Research Scientist at Meta. He was previously a Research Scientist at Google DeepMind. 

\textbf{Karim Jerbi} is Professor in the Psychology Department of the Université de Montréal and Associate Professor at Mila. He heads UNIQUE, the Quebec-wide Neuro-AI research center, and is also Canada Research Chair in Computational Neuroscience and Cognitive Neuroimaging. 

\textbf{Hamza Abdelhedi} is a Biomedical Engineering PhD student at the Université de Montréal, supervised by Karim Jerbi and specialising in Neuro-AI.

\textbf{Yorguin Mantilla Ramos} is a master's student at the Université de Montréal and Graduate Research Assistant at Mila. 

\textbf{Caglar Gulcehre} is Assistant Professor at the École Polytechnique Fédérale de Lausanne (EPFL) in Switzerland. He previously worked as a Staff Research Scientist at Google DeepMind on many topics in AI, including reinforcement learning, foundation models, novel architectures and training paradigms, safety + alignment, and natural language understanding.

\textbf{Mark Woolrich} is Professor of Computational Neuroscience at the University of Oxford. He is Head of Analysis and Associate Director of the Oxford centre for Human Brain Activity (OHBA). 

\textbf{Natalie Voets} is an Associate Professor at the Oxford Centre for Functional MRI of the Brain (FMRIB). She also works in the Awake Neurosurgery Service at the Oxford University Hospitals NHS Foundation Trust. Her research focuses on language mapping in surgical populations. %

\textbf{Oiwi Parker Jones} heads the \textit{Parker Jones Neural Processing Lab}
(PNPL\pnpl) in the Department of Engineering Science, University of Oxford. He is also a Fellow in Computer Science at Jesus College Oxford, a Principal Investigator at the Oxford Robotics Institute, and an Honorary Fellow in the Nuffield Department of Clinical Neurosciences.

%% file: sections/999_ack.tex
\begin{ack}

The authors would first like to thank our beta testers for feedback on the website, tutorial code, submission system, and leaderboard. %
We also gratefully acknowledge the use of the University of Oxford Advanced Research Computing (ARC) facility in carrying out this work (\url{http: //dx.doi.org/10.5281/zenodo.22558}), and NVIDIA for contributing additional GPUs used in this research. PNPL is supported by the MRC (MR/X00757X/1), Royal Society (RG\textbackslash
R1\textbackslash
241267), NSF (2314493), NFRF (NFRFT-2022-00241), and SSHRC (895-2023-1022). 

\end{ack}